\documentclass{article}
\usepackage{spconf,amsmath,graphicx}
\usepackage{times}
\usepackage{epsfig}
\usepackage{graphicx}
\usepackage{amsmath}
\usepackage{amssymb}
\usepackage{exscale}
\usepackage[mathscr]{eucal}
\usepackage{bm}
\usepackage{eqlist} 
\usepackage{flushend}
\usepackage{lettrine}
\usepackage{algorithm}
\usepackage{algorithmic}
\usepackage{url}
\usepackage{multirow} 
\usepackage{booktabs} 
\usepackage{colortbl} 
\usepackage{lscape}
\usepackage{times} 
\usepackage{amsfonts}
\usepackage{psfrag}
\usepackage{latexsym}
\usepackage{eucal}

\usepackage{multirow} 
\usepackage{booktabs} 
\usepackage{colortbl} 
\usepackage{algorithmic}
\usepackage{algorithm}
\usepackage{color}
\usepackage{xfrac}

\newcommand{\+}[1]{\bm{#1}}  
\newcommand{\argmin}{\operatornamewithlimits{argmin}}

\newcommand{\Balpha}{\bm{\alpha}} 

\title{Kernel Task-Driven Dictionary Learning for Hyperspectral Image Classification}
\name{Soheil Bahrampour$^{\dag}$ \qquad Nasser M. Nasrabadi$^{\ddag}$ \qquad Asok Ray$^{\dag}$  \qquad Kenneth W. Jenkins$^{\dag}$}

\address{$^{\dag}$ Pennsylvania State University, University Park, PA \\
    $^{\ddag}$ Army Research Laboratory, Adelphi, MD}
    
\begin{document}
%
\maketitle
\begin{abstract}
Dictionary learning algorithms have been successfully used in both reconstructive and discriminative tasks, where the input signal is represented by a linear combination of a few dictionary atoms. While these methods are usually developed under $\ell_1$ sparsity constrain (prior) in the input domain, recent studies have demonstrated the advantages of sparse representation using structured sparsity priors in the kernel domain. In this paper, we propose a supervised dictionary learning algorithm in the kernel domain for hyperspectral image classification. In the proposed formulation, the dictionary and classifier are obtained jointly for optimal classification performance. The supervised formulation is task-driven and provides learned features from the hyperspectral data that are well suited for the classification task. Moreover, the proposed algorithm uses a joint ($\ell_{12}$) sparsity prior to enforce collaboration among the neighboring pixels. The simulation results illustrate the efficiency of the proposed dictionary learning algorithm.
\end{abstract}
\section{Introduction}
\label{sec:intro}

Hyperspectral Imagery (HSI) has increasingly become popular for the remote sensing applications such as target detection~\cite{N14} and material identification~\cite{CTBA14}. Among several algorithms used for HSI classification~\cite{MB04, MCJ10, LBP10}, it has been shown that sparse representation classification (SRC) can achieve superior results~\cite{CNT11, YLYLT13}. For this purpose, a dictionary is usually constructed by collecting all the training samples, i.e. labeled pixels, and the underlying assumption is that the test pixel can be approximated with \textit{a few} dictionary atoms, i.e., test pixel lies in a low-dimensional subspace formed by the training samples that have the same label as the test pixel. However, the sparse coefficients generated by SRC can become unstable due to the high coherency of the dictionary atoms~\cite{IBP11}. This situation can be alleviated by enforcing similarity in the sparse codes of the neighboring pixels, which usually have similar spectral features, by an appropriate structured sparsity prior~\cite{LLYSYM13, MSMSDT14}. In particular, the joint sparsity prior assumes that the neighboring pixels lie in the same low-dimensional subspace. It enforces collaboration among these pixels and yields more stable sparse coefficients, which results in an improved classification performance~\cite{SQNT14}.

Recently, it has been shown that \textit{learning the dictionary}, rather than constructing it by using all the training samples, can significantly improve the performance of sparse representation-based algorithms for both reconstructive~\cite{WYGSM09} and discriminative tasks~\cite{MBZS08}. 
 Dictionary learning algorithms can generally be categorized into two groups: unsupervised and supervised methods. Unsupervised dictionary learning is aimed at finding a dictionary that yields the minimum errors for reconstruction tasks such as deniosing~\cite{EA06}, while supervised dictionary learning algorithms utilize the labels for minimizing a misclassification cost~\cite{MBZS08}. It has recently been shown that a task-driven formulation can achieve state-of-the-art performance in several classification tasks by jointly learning the dictionary and classifier~\cite{MBP12}.  

Similar to other machine learning methods, kernelized sparse representation algorithms which map the input into a higher-dimensional feature space using kernel function can result in significant performance improvements compared to the linear counterpart~\cite{NNT12, CNT13}. 
The rational is that when the data from different classes are projected into the kernel induced feature space, the classes become more separable and samples from the same classes can typically cluster together in subspaces resulting in more discriminative sparse codes. For this purpose, a few kernelized dictionary learning algorithms have been proposed~\cite{NPNC13, GGK12}. In~\cite{NPNC13}, an unsupervised learning is proposed by kernelizing the well-known K-SVD~\cite{AEB06} algorithm for object recognition. In~\cite{GGK12}, a supervised formulation has been proposed based on the Hilbert Schmidt independence criterion to maximize the dependency between the data and corresponding class labels. However, for a classification task, the preference is to utilize the labeled data to minimize a misclassification cost~\cite{MBP12}. 

In this paper, a kernelized task-driven dictionary learning algorithm is proposed in which a dictionary is trained to be optimal for HSI classification. The proposed algorithm generalizes the task-driven formulation of~\cite{MBP12} in two important ways. First, it enforces correlation among the neighboring pixels using the joint sparsity prior. Second, it generalizes the algorithm by providing a kernelized formulation. The proposed dictionary learning is obtained by solving a bi-level optimization problem which shows that, while the underlining joint sparse coding is non-smooth, the bi-level optimization cost is differentiable. The simulation results demonstrate that the proposed algorithm achieve state-of-the-art performance for HSI classification tasks.

\section{Background}
\label{sec:RelatedWorks}

\subsection{Dictionary learning}
\label{ssec:DicLea}
Dictionary learning has been widely used in various tasks such as reconstruction, classification, and compressive sensing~\cite{MBP12, ZL10, ZZH13a}. Let $\+X = \left[ \+x_{1}, \+x_{2}, \dots, \+x_{N} \right] \in \mathbb{R}^{n \times N}$ be the collection of $N$ (normalized) training HSI pixels where $n$ is the number of the spectral bands. In an unsupervised formulation, the dictionary $\+D \in \mathbb{R}^{n \times d}$ is usually obtained as the minimizer of the following cost~\cite{MBPS10}
\begin{equation} \label{eq:OnlineUnsupDicLea}
g\left(\+D\right) \triangleq \mathrm{E}_{\+x} \left[ l_u\left(\+x, \+D\right)\right],
\end{equation}
over the regularizing convex set $\mathcal{D} \triangleq \lbrace \+D \in \mathbb{R}^{n \times d}\vert \Vert \+d_{k} \Vert_{\ell_2} \leq 1, \forall k = 1, \dots, d \rbrace$, where $\+d_{k}$ is the $k^{th}$ column, or atom, in the dictionary and the unsupervised loss $l_u$ is defined as
\begin{equation}\label{eq:UnsupCost}
l_u\left(\+x, \+D\right) \triangleq \min_{\Balpha \in \mathbb{R}^{d}} \Vert \+x -\+D\Balpha \Vert_{2}^2 + \lambda_1\Vert \Balpha\Vert_{1} + \lambda_2\Vert \Balpha\Vert_{2}^2,
\end{equation}
which is the optimal value of the sparse coding problem with $\lambda_1$ and $\lambda_2$ being the regularizing parameters. It is assumed that the data $\+x$ is drawn from a finite probability distribution $p(\+x)$ which is usually unknown. A stationary point of the optimization problem can be efficiently obtained by an online optimization algorithm~\cite{MBPS10}.

The trained dictionary can then be used to (sparsely) reconstruct the inputs and the reconstruction error is usually a robust measure for classification tasks~\cite{BRNJ14, SMJMHJ14}. Other use of the trained dictionary is for feature learning where the sparse code $\Balpha^{\star}(\+x, \+D)$, obtained as a solution of~(\ref{eq:UnsupCost}), is used as input feature for training a classifier in the classical expected risk optimization framework~\cite{MBP12}. However, it has been shown that a more discriminative features can generally be obtained by learning the dictionary and classifier jointly in the following task-driven formulation~\cite{MBP12}
\begin{equation}\label{eq:taskDriv}
\min_{\+D \in \mathcal{D},\+W \in \mathcal{W}} \mathrm{E}_{\+y,\+x} \left[ l_{su}\left(\+y, \+W, \Balpha^{\star}(\+x, \+D)\right)\right] + \frac{\nu}{2} \Vert \+W \Vert_{F}^2,
\end{equation}
where $\+y \in \mathbb{R}^C$ is a binary vector representing the ground truth label of the input $\+x$ for a $C$-class classification problem, and $l_{su}$ is a (supervised) convex loss function that measures how well one can predict $\+y$ given the feature $\Balpha^{\star}$ and model parameters $\+W \in \mathcal{W}$, and $\nu$ is the regularizing parameter. In this paper, quadratic loss is used which is defined as
\begin{equation}\label{eq:QuadraticLoss}
l_{su}(\+y, \+W, \Balpha^{\star}) = \frac{1}{2}\Vert \+y - \+W\Balpha^{\star} \Vert_{\ell_2}^2, 
\end{equation}
and $\mathcal{W} = \mathbb{R}^{C \times d}$.

\subsection{Kernelized sparse representation with structured sparsity prior}
\label{ssec:SupDicLea}
Kernel methods are usually used to project the data set into a higher dimensional feature space to make different classes to become linearly separable. Let $\Phi: \mathbb{R}^n\rightarrow \mathcal{F}$ be a mapping from $\mathbb{R}^n$ to feature space $\mathcal{F}$ which can possibly be infinite-dimensional. It is assumed that $\mathcal{F}$ is a Hilbert space which allows the use of Mercer kernels to carry out the projection implicitly. Mercer kernel $\operatorname{k}(\+x_1,\+x_2): \mathbb{R}^n\times \mathbb{R}^n\rightarrow \mathcal{R}$ is a function defined as $\operatorname{k}(\+x_1,\+x_2) = <\Phi(\+x_1), \Phi(\+x_2)>$ where $<>$ is the inner product operator~\cite{CV95}. Among commonly used kernel functions are the Gaussian kernel $\operatorname{k}(\+x_1,\+x_2) =\operatorname{exp}\left( -\frac{\Vert \+x_1-\+x_2 \Vert^2}{\sigma}\right) $ and polynomial kernel $\operatorname{k}(\+x_1,\+x_2) =(<\+x_1, \+x_2>)^c$, where $\sigma$ and $c$ are the kernel parameters.

The kernel sparse representation of the input feature $\Phi(\+x)$ can then be obtained by solving~\cite{NPNC13}
\begin{equation*}\label{eq:KernelSparse}
 \min_{\Balpha \in \mathbb{R}^{d}} \Vert \Phi(\+x) -\Phi(\+D)\Balpha \Vert_{2}^2 + \lambda_1\Vert \Balpha\Vert_{1} + \lambda_2\Vert \Balpha\Vert_{2}^2,
\end{equation*} 
where $\Phi(\+D) = \left[ \Phi(\+d_1) \dots \Phi(\+d_N)\right] $ and $\+d_j$ are the columns of $\+D$. Note that $\Vert \Phi(\+x) -\Phi(\+D)\Balpha \Vert_{2}^2 = \operatorname{k}(\+x,\+x)-2\Balpha^T\operatorname{k}(\+D,\+x)+\Balpha^T\operatorname{k}(\+D,\+D)\Balpha$ and no explicit mapping into the feature space is required to solve the optimization problem. As discussed in previous section, the neighboring HSI pixels usually have similar spectral features and more robust sparse codes can be obtained if they are jointly reconstructed~\cite{SQNT14, CNT13}. Let $\lbrace\+x^1,\dots, \+x^S\rbrace$ be the set of $S$ neighboring pixels centered at $\+x^1$ which are denoted as $\lbrace\+x^s\rbrace$ in this paper. Joint sparsity enforces the neighboring pixels to be represented in the same subspace and the optimal sparse coefficients $\+A^{\star}(\lbrace\+x^s\rbrace,\+D)$ are obtained by solving following optimization problem
\begin{equation} \label{eq:KernJSRC}
\argmin_{\+A \in \mathbb{R}^{d\times S}} \frac{1}{2}\sum_{s=1}^S\Vert \Phi(\+x^s) - \Phi(\+D)\Balpha^s\Vert_{2}^2 +\lambda_1\Vert \+A \Vert_{\ell_{12}} + \frac{\lambda_2}{2}\Vert \+A \Vert_{F}^2, 
\end{equation}
where $\Balpha^s$ is the sparse code for pixel $\+x^s$ and $\Vert \+A \Vert_{12} = \sum_{j=1}^d \Vert \+a_{j\rightarrow} \Vert_{2}$ in which $\+a_{j\rightarrow}$'s are the rows of $\+A$. The above optimization problem encourages row sparsity in $\+A^\star$ and therefore the neighboring pixels are enforced to be jointly reconstructed by the same sparse code pattern~\cite{SQNT14}. 

\section{Kernelized task-driven dictionary learning}
\label{sec:MulModDicLea}
This section extends the task-driven dictionary learning algorithm by using joint sparsity prior, which enforces collaboration among the neighboring HSI pixels. Moreover, we extend the algorithm to the kernel domain which provides a general framework for task-driven dictionary learning using arbitrary kernel functions. With the same notations from previous section, and without loss of generality, let the input signal consist of $S$ neighboring pixels $\lbrace\+x^s\rbrace$ centered at $\+x^1$ and the label vector of the center pixel be $\+y$ . We propose to obtain the dictionary $\+D^\star$ and the model parameter $\+W^\star$ jointly in the kernel space as the minimizer of the following optimization
\begin{equation}\label{eq:kernTaskDriv}
\min_{\+D \in \mathcal{D}, \+W \in \mathcal{W}} \mathrm{E} \left[ l_{su}(\+y, \+W, {\Balpha^\star}^1(\lbrace\+x^s\rbrace,\+D))\right] + \frac{\nu}{2} \Vert \+W \Vert_{F}^2,
\end{equation}
where ${\Balpha^\star}^1$ is the first column of the minimizer $\+A^{\star}(\lbrace\+x^s,\+D^s\rbrace )$ of the optimization problem~(\ref{eq:KernJSRC}), which is the sparse code for the center pixel, and $l_{su}$ is defined in Eq.~(\ref{eq:QuadraticLoss}). It should be noted that while $l_{su}$ is chosen to be the quadratic loss for simplicity, the formulation can be easily extended to any other convex cost functions such as those used in~\cite{MBP12}. The expectation is taken with respect to the joint probability distribution of the HSI inputs $\lbrace\+x^s\rbrace$ and label $\+y$. 

The main difficulty in optimizing~(\ref{eq:kernTaskDriv}) is the nondifferentiability of $\+A^{\star}(\lbrace\+x^s,\+D^s\rbrace )$. However, it can be shown that the sparse coefficients $\+A^{\star}$ is  differentiable almost everywhere. To prove that, one can use the optimality condition of $\+A^{\star}$
\begin{equation}\label{eq:OptimalityCond}
\left\lbrace \begin{array}{l}\begin{aligned}
&\left[\operatorname{k}({\+d_j},\+x^1) \dots \operatorname{k}({\+d_j},\+x^S) \right]-  \operatorname{k}({\+d_j},\+D) {\+A}^{\star} \\ 
&-\lambda_2 \+a_{j\rightarrow}^{\star} = \lambda_1\frac{\+a_{j\rightarrow}^{\star}}{\Vert \+a_{j\rightarrow}^{\star} \Vert_{\ell_2}}, \textrm{ if } \Vert \+a_{j\rightarrow}^{\star} \Vert_{\ell_2} \neq 0, \end{aligned}\\
\begin{aligned}
&\Vert \left[\operatorname{k}({\+d_j},\+x^1) \dots \operatorname{k}({\+d_j},\+x^S) \right]-  \operatorname{k}({\+d_j},\+D) {\+A}^{\star} \\
&-\lambda_2 \+a_{j\rightarrow}^{\star} \Vert_{\ell_2} \leq \lambda_1, \textrm{otherwise},
\end{aligned}
\end{array}\right.
\end{equation}
which is obtained by subgradient of the cost function. For the solution ${\+A}^{\star}$, the active set is defined to be
\begin{equation}\label{eq:ActSet}
\Lambda = \lbrace j \in \lbrace 1, \dots, d\rbrace: \Vert \+a_{j\rightarrow}^{\star} \Vert_{\ell_2} \neq 0 \rbrace,
\end{equation}
where $\+a_{j\rightarrow}^{\star}$ is the $j^{th}$ row of $\+A^{\star}$. It can be shown that the active set is locally constant for the small perturbation of $\lbrace\+x^s\rbrace, \+D$ and, therefore, $\+A^{\star}$ is locally differentiated. Moreover, similar to the procedure in~\cite{MBP12, BNRJ15a},  it can be shown that the set of points where the active set changes has measure zero and therefore $\mathrm{E}\left[ l_{su}\left(\+y, \+W, {\Balpha^\star}^1\right) \right]$ is differentiable on $\mathcal{D}\times\mathcal{W}$, and the gradients can be computed using chain rule. The detailed proof is a bit involved and is omitted here due to the space limitation. The algorithm to find the optimal dictionary ${\+D}$ and model parameter $\+W^\star$ for HSI classification is described in Algorithm~\ref{alg:KerTaskDriDic}. In the special case when $S=1$ and linear kernel is chosen, the proposed algorithm reduces to the task-driven dictionary learning algorithm in~\cite{MBP12}. In theory, one needs to select $\lambda_2$ in Eq.~(\ref{eq:KernJSRC}) to be strictly positive which guarantees the linear equation in the algorithm (step 7) to have unique solution. In other words it is easy to show that the matrix $( \operatorname{k}(\+{D}_{\Lambda},\+{D}_{\Lambda})\otimes\+I + \lambda_1\+{\Delta} + \lambda_2\+{I})$ in Algorithm~\ref{alg:KerTaskDriDic} is positive definite given $\lambda_1 \geq 0, \lambda_2 > 0$. However, in practice it is observed that setting $\lambda_2$ to zero yields satisfactory results. As in any nonconvex optimization problem, if the algorithm is not initialized properly, it may yield poor performance. In this paper, we used unsupervised dictionary learning with stochastic gradient descent to initialize $\+D$. Once dictionary $\+D$ is initialized, the initial value of $\+W$ is set by solving~(\ref{eq:taskDriv}) only with respect to $\+W$ which is a convex optimization problem. 

\begin{algorithm}[!t]
\footnotesize 
\caption{\small Stochastic gradient descent algorithm for the kernelized task-driven dictionary learning under the joint sparisty prior}\label{alg:KerTaskDriDic}
\begin{algorithmic}[1]
	\REQUIRE Kernel function $\operatorname{k}$, neighborhood size $S$, Regularization parameters $\lambda_1, \lambda_2, \nu$, learning rate parameters $\rho, t_0$, number of iterations $T$, initial dictionary $\+D \in \mathcal{D}$, and initial model parameter $\+W \in \mathcal{W}$.
	\ENSURE Learned $\+D$ and $\+W$ 
        \FOR{$t= 1, \dots, T$}
        	\STATE Draw a sample $(\+x^1, \dots, \+x^S, {\+y})$ where $\+x^1$ is a training pixel randomly selected from the training set with label ${\+y}$ and $(\+x^2, \dots, \+x^S)$ are its closest $(S-1)$ HSI pixels.
        	\STATE Find solution $\+A^{\star} = \left[{\Balpha^\star}^1 \dots {\Balpha^\star}^S\right] = \left[{\+a_{1\rightarrow}^{\star}}^T \dots  {\+a_{d\rightarrow}^{\star}}^T\right]^T \in \mathbb{R}^{d\times S}$ of the joint sparse coding problem~(\ref{eq:KernJSRC}).
        	\STATE Compute the set of active rows $\Lambda$ of $\+A^{\star}$ using~(\ref{eq:ActSet}).
        	\STATE Let $\+{D}_{\Lambda} \in \mathbb{R}^{n\times |\Lambda|}$ and $\+{W}_{\Lambda} \in \mathbb{R}^{C\times |\Lambda|}$ be formed by the columns of $\+{D}$ and $\+{W}$ which are indexed in $\Lambda$.
        	\STATE Compute $\+{\Delta} = \+{\Delta}_1\oplus \dots \oplus \+{\Delta}_{|\Lambda |} \in \mathbb{R}^{S|\Lambda| \times S|\Lambda|}$, where $\+{\Delta}_j = \frac{1}{\Vert \+a_{j\rightarrow}^{\star} \Vert_{\ell_2}}\+I - \frac{1}{{\Vert \+a_{j\rightarrow}^{\star} \Vert_{\ell_2}}^3}{\+a_{j\rightarrow}^{\star}}^T\+a_{j\rightarrow}^{\star} \in \mathbb{R}^{S \times S}, \forall j \in \Lambda$, $\+I$ is the identity matrix, and $\oplus$ is the direct sum operator. 
        	\STATE Compute $\+{\beta} \in \mathbb{R}^{dS}$ as:
\begin{equation*}
\+{\beta}_{\Upsilon^c} = \+0, \+{\beta}_{\Upsilon} = ( \operatorname{k}(\+{D}_{\Lambda},\+{D}_{\Lambda})\otimes\+I + \lambda_1\+{\Delta} + \lambda_2\+{I})^{-1}\+g,
\end{equation*}
where $\Upsilon = \cup_{j \in \Lambda} \lbrace j, j+d, \dots, j+(S-1)d\rbrace $, $\+{\beta}_{\Upsilon}$ is a vector in $\mathbb{R}^{|\Upsilon|}$ whose rows are those of $\+{\beta}$ indexed by $\Upsilon$,  $\otimes$ is the Kronecker product, $\+g = \operatorname{vec}\left( (\+W\bar{\+{A}}-\bar{\+{Y}})^T \+{W}_{\Lambda}\right) $, $\bar{\+{A}}=\left[{\Balpha^\star}^1, \+0, \dots, \+0 \right] \in \mathbb{R}^{d\times S}$, $\bar{\+{Y}}=\left[{\+y}^{}, \+0, \dots \+0, \right] \in \mathbb{R}^{C\times S}$, and $\operatorname{vec}(.)$ is the vectorization operator.
        	\STATE Choose the learning rate $\rho^{}_t \leftarrow \min(\rho, \rho\frac{t_0}{t})$.
        	\STATE Update the parameters by a projected gradient step:
        	    \begin{equation*}
        		\begin{aligned}
&\+W\leftarrow \+W -\rho^{}_t\left( (\+W{\Balpha^\star}^1-{\+y}) {{\Balpha^\star}^1}^T + \nu \+W \right) , \\
& \begin{split} \+D\leftarrow \Pi_{\mathcal{D}}\left[ \+D - \rho^{}_t\sum_{s=1}^S \right. & \Big( \left[\operatorname{k}^{\prime}(\+x^s,\+d^{}_1)-\operatorname{k}^{\prime}(\+{D},\+d^{}_1){\Balpha^s}^\star \dots  \right.\\
& \left. \operatorname{k}^{\prime}(\+x^s,\+d^{}_d)-\operatorname{k}^{\prime}(\+{D},\+d^{}_d){\Balpha^s}^\star \right]\operatorname{diag}(\+{\beta}_{\tilde{s}})  \\
&-\left[ \operatorname{k}^{\prime}(\+{D},\+d^{}_1)\+{\beta}_{\tilde{s}}{\alpha^s_1}^\star\dots \operatorname{k}^{\prime}(\+{D},\+d^{}_d)\+{\beta}_{\tilde{s}}{\alpha^s_d}^\star\right] \Big) \Bigg], 
\end{split}
\end{aligned}
\end{equation*}
where $\tilde{s} = \left\lbrace s, s+ S, \dots, s+(d-1)S\right\rbrace $ and $\operatorname{k}^{\prime}(\+{D},\+d^{}_k) = \left[\frac{\partial \operatorname{k}(\+d^{}_1,\+d^{}_k)}{\partial \+d^{}_k} \dots \frac{\partial \operatorname{k}(\+d^{}_d,\+d^{}_k)}{\partial \+d^{}_k} \right] \in \mathbb{R}^{n \times d}$.
        \ENDFOR
\end{algorithmic}
\end{algorithm}

\section{Results and discussion}
\label{sec:Results}


\begin{table*}[!t] 
\caption{Average and overall accuracy obtained for HSI classification of the Indian Pine image. }
\label{tab:Results1}
\centering
\small
\setlength{\tabcolsep}{2mm}
\begin{tabular}{ccccccccccc}
 & SVM-l & SVM-k & SRC-$\ell_1$-l & SRC-$\ell_1$-k  & SRC-$\ell_{12}$-l & SRC-$\ell_{12}$-k  & SDL-$\ell_1$-l & SDL-$\ell_1$-k & SDL-$\ell_{12}$-l & SDL-$\ell_{12}$-k\\
\toprule
 & & & \multicolumn{4}{c}{Dictionary size $d=997$} &  \multicolumn{4}{c}{Dictionary size $d=80$}\\
 \cmidrule(lr){4-7} \cmidrule(lr){8-11} 
OA&64.94&75.78&71.88&74.83&{76.41}&{{77.41}}&{81.43}&{{83.48}}&84.14&\textbf{87.56}  \\
AA&56.53&61.40&64.28&67.19&{64.67}&{{63.66}}&{66.43}&{{74.65}}&{76.56}&{{\textbf{81.25}}} \\
\bottomrule
\end{tabular}
\end{table*} 

\begin{table*}[!t] 
\caption{Average and overall accuracy obtained for HSI classification of the University of Pavia image.}
\label{tab:Results2}
\centering
\small
\setlength{\tabcolsep}{2mm}
\begin{tabular}{ccccccccccc}
& SVM-l & SVM-k & SRC-$\ell_1$-l & SRC-$\ell_1$-k  & SRC-$\ell_{12}$-l & SRC-$\ell_{12}$-k  & SDL-$\ell_1$-l & SDL-$\ell_1$-k & SDL-$\ell_{12}$-l & SDL-$\ell_{12}$-k\\
\toprule
 & & & \multicolumn{4}{c}{Dictionary size $d=3921$} &  \multicolumn{4}{c}{Dictionary size $d=45$}\\
 \cmidrule(lr){4-7} \cmidrule(lr){8-11} 
OA& 61.84&62.43&{66.51}&{74.05}&{83.86}&{{82.67}}&{69.30}&{{81.25}}&{84.48}& \textbf{86.07} \\
AA& 65.09&72.14&{75.98}&{80.06}&{86.29}&{{85.28}}&{83.44}&{{82.24}}&{84.47}&{{\textbf{87.37}}}\\
\bottomrule
\end{tabular}
\end{table*} 

The performance of the proposed HSI classification algorithm is evaluated on the Indian Pine image, which is generated by Airborne Visible/Infrared Imaging Spectrometer (AVIRIS), and the University of Pavia image. The Indian Pine image contains 16 classes spread over the $145 \times 145$ pixels and each pixel has 220 bands ranging from $0.2$ to $2.4\mu m$. The 20 bands corresponding to the water absorption are removed before processing the image. Similar to the setup in~\cite{SQNT14}, we randomly select 997 pixels ($10.64 \%$ of the available data) to form the training set and the rest of the pixels are used for testing. The University of Pavia image is an urban image and has 115 spectral bands ranging from $0.43$ to $0.86\mu m$. It contains 9 classes spread over the $610 \times 340$ pixels. The 12 noisiest bands are removed. For this dataset, the standard training and test split is used~\cite{SQNT14} where the training set consists of $3,921$ pixels ($10.64\%$ of the available data) and the rest $40,002$ pixels are used for testing. For the dictionary learning algorithms, the size of the dictionary is chosen to be 5 atoms per class. The regularization parameters $\lambda_1$ and $\nu$ and Gaussian kernel parameter $\sigma$ are selected using cross-validation on the sets $\lbrace0.001, 0.01, 0.1\rbrace$, $\lbrace10^{-8}, 10^{-7}, \dots, 10^{-1}\rbrace$, and $\lbrace 0.5, 1, \dots, 5\rbrace$, respectively, and $\lambda_2$ is set to zero. The learning parameters $\rho$ and $t_0$ are selected similar to the procedure outlined in~\cite{MBP12}. 
 
The performance of the proposed kernelized dictionary learning algorithm is compared with the linear task-driven dictionary learning algorithm (SDL-$\ell_1$-l) proposed in~\cite{MBP12}. For this purpose, we report the results of our proposed algorithm using three different settings which are named as SDL-$\ell_1$-k, SDL-$\ell_{12}$-l, SDL-$\ell_{12}$-k. The SDL-$\ell_1$-k is the extension of the SDL-$\ell_1$-l to the kernel domain. The SDL-$\ell_{12}$-l is the enforcing collaboration of the neighboring pixel using the joint sparsity and in the linear domain. Finally,  the SDL-$\ell_{12}$-k is the setting where the neighboring pixels are jointly reconstructed in the kernel domain. We also evaluate the performance of the proposed algorithm against linear and kernel SVM,  namely SVM-l and SVM-k respectively, as well as the sparse-based representation classification algorithms. For the latter, all the training samples are used to construct the dictionary and the results are reported using $\ell_1$ and $\ell_{1,2}$ priors in both linear and kernel domains which are named as SRC-$\ell_1$-l, SRC-$\ell_1$-k, SRC-$\ell_{12}$-l, and SRC-$\ell_{12}$-k, accordingly.

The classification results on the Indian Pine and University of Pavia hyperspectral Images are shown in Table~\ref{tab:Results1} and Table~\ref{tab:Results2}, respectively. As expected, the kernelized formulations usually achieve better classification performance. Moreover, it is consistently observed that using joint sparsity prior ($\ell_{12}$ norm) to enforce collaboration among the neighboring pixels improves the performance. The proposed SDL-$\ell_{12}$-k achieves the best performance against the competitive algorithms for both datasets. In comparing the performances of the dictionary-learning based algorithms with those in which the dictionary is constructed by collecting all the training samples, one should also note the difference in the dictionary sizes. The proposed task-driven formulations achieve the better performances with more compact dictionaries which translates into more computationally efficient processing of the test samples.

\section{Conclusions}
\label{sec:Conclusions}
In this paper, a kernelized task-driven dictionary learning algorithm is proposed for supervised HSI classification. The proposed formulation enjoys a joint sparsity prior which enforces collaboration among the neighboring pixels for robust sparse representation. It is shown that the proposed algorithm, equipped with compact dictionary, achieves state-of-the-art performances for classification of the Indian Pine and the University of Pavia hyperspectral images. The proposed formulation provides a general framework for nonlinear supervised dictionary learning that can be readily applied to other classification tasks. Future research topics includes extension of the proposed algorithm to include other structured sparsity priors and testing them on different classification tasks. 

\bibliographystyle{IEEEbib}
\bibliography{referencesICASSP}

\end{document}